\documentclass[runningheads,a4paper]{llncs}

\usepackage{amssymb}
\usepackage{graphicx}

\usepackage{url}
\urldef{\mailsa}\path|{alfred.hofmann, ursula.barth, ingrid.haas, frank.holzwarth,|
\urldef{\mailsb}\path|anna.kramer, leonie.kunz, christine.reiss, nicole.sator,|
\urldef{\mailsc}\path|erika.siebert-cole, peter.strasser, lncs}@springer.com|    
\newcommand{\keywords}[1]{\par\addvspace\baselineskip
\noindent\keywordname\enspace\ignorespaces#1}

\begin{document}

\mainmatter  % start of an individual contribution

\title{Iterative Visual Recognition for Learning Based Randomized Bin-Picking}

\titlerunning{Iterative Visual Recognition}

\author{Kensuke Harada$^{2,1}$ \and Weiwei Wan$^1$ \and Tokuo Tsuji$^{3,1}$ \and Kohei Kikuchi$^{4}$ \and \\
Kazuyuki Nagata$^1$ \and Hiromu Onda$^1$}
\authorrunning{K. Harada et al.}

\institute{
$1$: National Inst. of Advanced Industrial Science and Technology, Tsukuba, Japan, \\
$2$: Osaka Univ., Toyonaka, Japan\\
$3$: Kanazawa Univ., Kanazawa, Japan \\
$4$: Toyota Motors Co. Ltd., Toyota, Japan}

\toctitle{}
\tocauthor{}
\maketitle

\begin{abstract}
This paper proposes a iterative visual recognition system for learning based randomized bin-picking. 
%For the purpose of increasing the performance of our bin-picking method, 
%we need maximum visibility of randomly stacked objects to detect the pose of a number of objects. 
%To effectively detect the pose of stacked objects, 
%we consider using the previous information on the object pose detection. 
Since the configuration on randomly stacked objects while executing the current picking trial is just partially 
different from the configuration while executing the previous picking trial, 
we consider detecting the poses of objects just by using a part of visual image taken at the current picking trial 
where it is different from the visual image taken at the previous picking trial. 
By using this method, we do not need to try to detect the poses of all objects included in the pile 
at every picking trial. 

Assuming the 3D vision sensor attached at the wrist of a manipulator, we first 
explain a method to determine the pose of a 3D vision sensor maximizing the visibility of randomly stacked objects. 
Then, we explain a method for detecting the poses of randomly stacked objects. 
Effectiveness of our proposed approach is confirmed by experiments using a dual-arm manipulator where a 3D vision sensor and 
the two-fingered hand attached at the right and the left wrists, respectively. 

\end{abstract}

\section{Introduction}

Randomized bin-picking refers to the problem of automatically picking 
an object that is randomly stored in a box. If randomized bin-picking is introduced to a production process, 
we do not need any parts-feeding machines or human workers 
to once arrange the objects to be picked by a robot. Although a number of researches have been done on 
randomized bin-picking \cite{Domae2014,SII,Dupuis_08,icra13}, 
randomized bin-picking is still difficult and is not widely introduced to production processes. 
Since one of the main reasons is its low success rate of the pick, we have proposed a learning based approach 
which can automatically increase the success rate \cite{RA-L}. 
%To estimates whether or not the pick will success, 
%a discriminator is trained through a number of picking trials based on 
%relation between the visual information on the pile and the result of a picking trial. 

\begin{figure}[t]
	\centering
	\includegraphics[width=6cm]{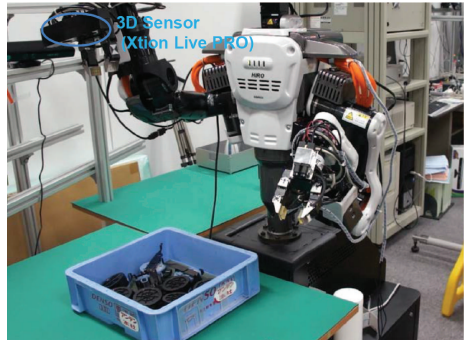}
	\caption{Overview of our bin-picking system \label{intro:RA-L2015}}
\end{figure}

Fig. \ref{intro:RA-L2015} illustrates the randomized bin-picking where 
we use a dual-arm manipulator with a vision sensor (3D depth sensor) and two-fingered grippers both attached at the wrist. 
We first detect the poses of randomly stacked objects by using the visual information obtained 
from the 3D vision sensor attached at the wrist. Once the objects' poses are obtained, we consider 
predicting whether the robot can successfully pick one of the objects from the pile. 
If it is predicted that the robot successfully picks an object, the robot tries to pick an object. 
In our approach, the success rate is expected to increase if the number of detected object increases. 
Here, in the conventional research on randomized bin-picking, we have tried to detect the poses of all objects 
at every picking experiment in spite of the fact that the configuration of object while executing the current picking 
trial is almost same as that while executing the previous picking trial. 
The configuration on objects while executing the current picking trial is usually just partially different from 
that while executing the previous picking trial since a finger usually contacts just a few objects 
during the previous picking trial. 
To cope with this problem, we propose a new method for object pose detection for the randomized bin-picking. 
In our proposed method, we consider detecting the objects' poses at a portion of the pile where its visual 
information is different from the visual information obtained during the previous picking experiment. 

In our proposed method, we first obtain the pose of 3D vision sensor attached at the wrist 
to capture the point cloud on the randomly stacked pile realizing its maximum visibility. 
Here, based on the occupancy grid map, the maximum visibility of the pile is realized 
by merging the point cloud captured during the current picking trial with the point cloud captured 
during the previous picking trial. Then, we show a method for detecting the poses of objects. 
We consider comparing each segment of the point cloud captured during current picking trial with that captured 
during the previously picking trial. If the difference is small, we do not estimate the poses of objects 
and can save the time needed for the estimation. 

\section{Learning Based Bin-Picking Overview}

\begin{figure}
	\centering
	\includegraphics[width=9cm]{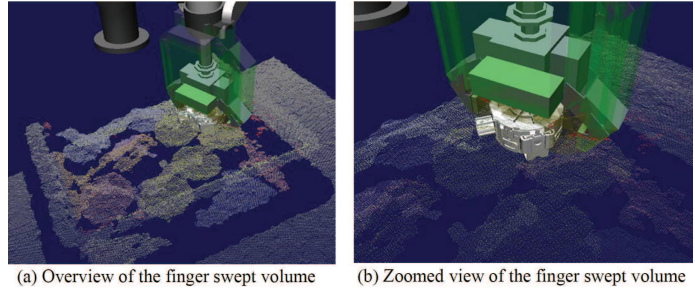}
	\caption{Finger swept volume \label{fig:sweepvol}}
\end{figure}

%We first explain the method of randomized bin-picking used in our research. 
We first briefly explain the leaning based bin-picking proposed previously \cite{RA-L}. 
As shown in Fig. \ref{intro:RA-L2015}, let us consider the case in which the same objects are randomly stored in a box. 
%By using a two-fingered gripper attached at the tip of a manipulator, we consider performing the randomize bin-picking. 
To pick an object from the pile, a 3D vision sensor (e.g., Xtion PRO) first captures a point cloud of randomly stacked objects. 
%Then, we segment the captured point cloud. For each segment of point cloud which bounding-box size is similar to the bounding-box size of an object, 
Then, we try to estimate the poses of randomly stacked objects. 
Then, we try to pick one of the objects which poses were detected. 
First among multiple candidates of grasping postures,  we solve IK to check the reachability of the robot. 
Then, for each reachable grasping posture, a discriminator trained through a number of picking trials 
estimates whether or not the robot can successfully pick an object. Here, the estimation is performed based 
on the distribution of point cloud included in the swept volume of finger motion as shown in Fig. \ref{fig:sweepvol}. 
If there are multiple grasping postures which are estimated to successfully pick an object, 
we consider selecting a grasping posture from multiple candidates according to the value of an index function. 
Then, the robot actually picks an object according to the selected grasping posture. 
%This research defines that the pick is successful if the gripper grasps the target object, lifts it up, 
%and places it out of the box. 

\section{Sensor Pose Calculation}

\begin{figure}[thb]
\centering
  \includegraphics[width=6cm]{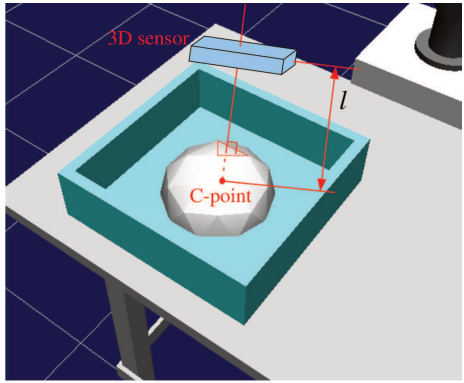}
  \caption{Regular polygon assumed at the geometrical center of bottom surface \label{fig:polygon}}
\end{figure}

We assume that the manipulator has at least 6 DOF such that the wrist can make an arbitrary pose within its movable range. 
The pose of the 3D vision sensor is determined to maximize the visibility of 
randomly stacked objects so as to precisely estimate the poses of randomly stacked objects. 
As shown in Fig. \ref{fig:polygon}, let us assume a $n$-faced regular polyhedron sharing its geometrical center 
with the geometrical center of box's bottom surface. 
Let us also assume a line orthogonally intersecting a face of the polyhedron 
and passing through the geometrical center. Let us consider a point along the line where the distance measured 
from the geometrical center is $l$. We make a 3D vision sensor locating at this point and facing the geometrical center. 
By discretizing a position of a 3D vision sensor along the line as $l=l_1, l_2, \cdots, l_m$, 
we can totally assume $m \cdot n$ candidates of a 3D sensor's pose. 
We consider imposing the following conditions for each candidate: 

\begin{itemize}
\item[] The 3D vision sensor is located above the box's bottom surface. 
\item[] IK (inverse kinematics) of the arm where the 3D vision sensor is attached at its wrist is solvable. 
\item[] For a pose of the 3D vision sensor where IK is solvable, no collision occurs among the links 
and between a link and the environment. 
\end{itemize}

\begin{figure}[thb]
\centering
  \includegraphics[width=12cm]{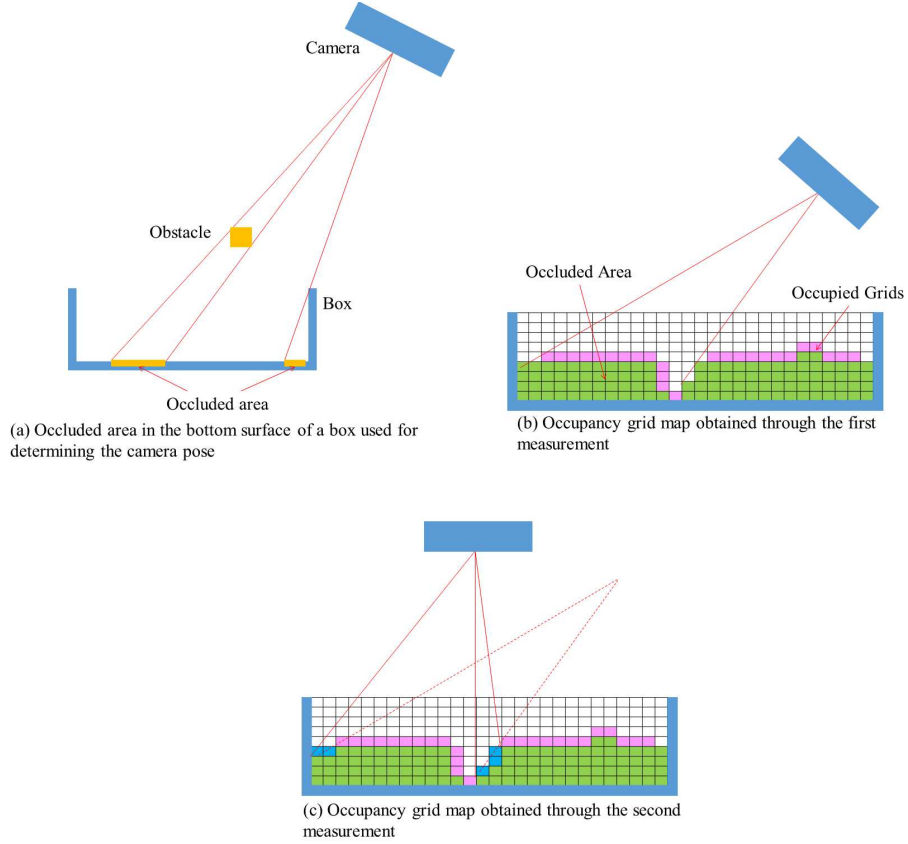}
  \caption{Determination of camera pose maximizing the visibility of stacked objects \label{fig:occupancy}}
\end{figure}

Among a set of 3D sensor's pose satisfying the above conditions, we consider selecting one 
maximizing the visibility of randomly stacked objects. 
Here, robotic bin-picking is usually iterated until there is no object remained in a box. 
For the first picking trial, we consider selecting a 3D sensor's pose minimizing the 
occluded area of the box's bottom surface as shown in Fig. \ref{fig:occupancy} (a). 
After the second picking trial, we consider using previous result of measurement to determine 
the pose of a 3D sensor as shown in Fig. \ref{fig:occupancy} (b) and (c). We consider partitioning the 
storage area into multiple grid cells \cite{Thrun,Nagata10}. By using the point cloud captured in the 
previous picking experiment, we mark {\it occupied} to the grid cells including the 
point cloud. We also mark {\it occluded} to the grid cells occluded by the grid cells marked as {\it occupied}. 
Pose of a 3D sensor is determined to maximize the number of grid cells marked as {\it occluded} to be visible. 

Here, through the previous picking experiment, configuration of stacked objects may change since the manipulator 
contacts the objects. However, the method explained in this subsection does not consider the change of configuration. 
Our method approximates the optimum pose of the 3D sensor by assuming the change of configuration is small. 

\section{Object Pose Detection}

This section explains a method for detecting the pose of randomly stacked objects. 
For the first picking trial, we consider detecting the poses of as many objects as possible. 
After the second picking trial, we consider detecting the poses of objects which poses are changed. 

\subsection{Object Pose Detection for the First Picking Trial}

\begin{figure}
	\centering
	\includegraphics[width=12.cm]{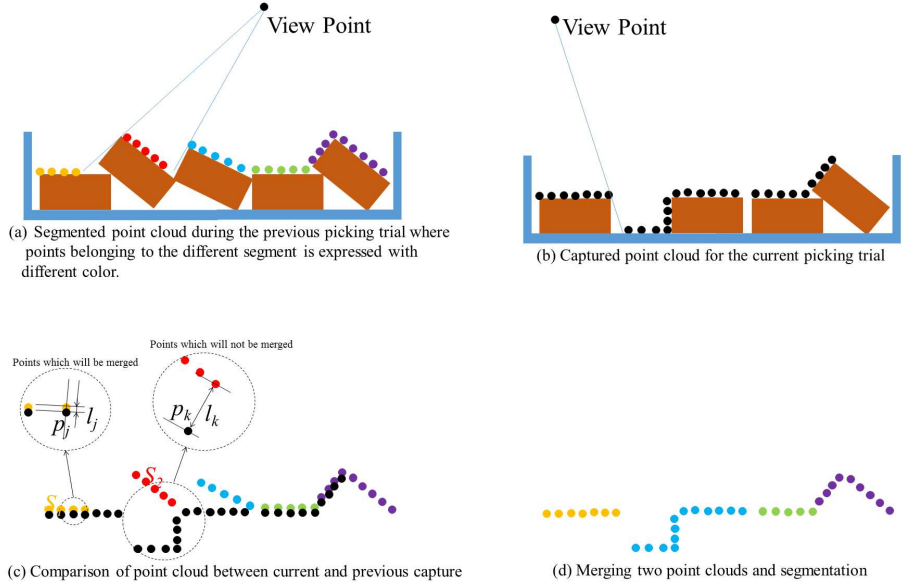}
	\caption{Segmentation of point cloud after the second picking trial \label{fig:merge}}
\end{figure}

To pick an object from the pile, the 3D vision sensor first 
captures a point cloud of randomly stored objects. Then, we segment the captured point cloud as shown in Fig. \ref{fig:merge}(a). 
In this research, we used a segmentation method based on the Euclidian cluster prepared in the PCL (Point Cloud Library) \cite{pcl}. 
%Next, we consider comparing the bounding-box size of each segment with that of the object model. 
For each segment of point cloud which bounding-box size is similar to the bounding-box size of an object, we try to 
estimate the pose of an object using a two-step algorithm: 
first roughly detecting the pose by using the 
CVFH (Clustered Viewpoint Feature Histogram) \cite{CVFH} and the CRH (Camera Roll Histogram) 
estimation, and then detecting the precise pose by using the ICP (Iterative Closest Point) estimation method. 
In a preprocessing process before starting the detection, we prepared 42 
partial view of the object model, and precompute the CVFH 
and CRH features of each view. During the detection, we extract the 
plenary surface from the point cloud, segment the remaining points cloud, and
compute the CVFH and CRH features of each segmentation. Then, we match the 
precomputed features with the features of each segment and estimate the orientation
of the segmentations. 
The matched segments are further refined using ICP estimation method to ensure good matching. 
The segmentation that has highest ICP matches and smallest outlier points will be
used as the output. 

For the first picking trial, we usually detect the poses of a number of objects. 
In such cases, since we have to solve ICP estimation for a number of times, we consider using 
multiple threads and solving multiple ICP estimation in parallel. 

\subsection{Object Pose Detection after the Second Picking Trial}

After the second picking trial, we consider using the current point cloud 
together with the previously captured one. 
%We consider comparing the currently captured point cloud with the previously captured one. 
If a part of the previously captured point cloud is similar to the current one, 
we do not need to calculate the object's pose belonging to the part of point cloud and can save 
the time needed to calculate the objects' poses. 
%Also, by adequately locating a 3D sensor, we can obtain a point cloud of randomly stacked 
%objects with less occluded area since we can merge the previously captured point cloud to the current one. 
In a picking task, after a 3D sensor capture a point cloud of randomly stacked objects, a 
robot manipulator tries to pick an object from the pile. The configuration of objects after a robot manipulator tries to pick an object 
is usually partially different from the configuration before the picking trial. 
If the previously captured point cloud is partially similar to the current point cloud, we consider 
merging the part of previously captured point cloud to the current one. 
By merging a part of the previous point cloud to the current one, the occluded area of the point cloud 
is expected to be smaller. 

The algorithm of merging the point cloud is outlined in Fig. \ref{fig:merge} and Algorithm 1. 
Let $\bar{P} = (\bar{p}_1, \bar{p}_2, \cdots, \bar{p}_m)$ and 
$P = (p_1, p_2, \cdots, p_n)$ be the previously captured point cloud and the current one, respectively. 
Let also $\bar{P}_1, \bar{P}_2, \cdots$, and $\bar{P}_s$ be the segments of previous point cloud. 
The overview of the merging algorithm is explained in the following. 
Fig. \ref{fig:merge} (a) shows the segmented point cloud obtained during the previous picking trial. 
On the other hand, Fig. \ref{fig:merge} (b) shows the current point cloud where 
the configuration of object is partially different from the previous one. 
As shown in Fig. \ref{fig:merge} (c), for each point included in the current point cloud, 
we search for the point included in the previous point cloud making the minimum distance between them (lines 6 and 7). 
We further find a segment of previous point cloud where the point making the minimum distance belongs to (line 8). 
For each segment of previous point cloud, we introduce two integer numbers ${\rm near}(i)$ and {\rm far}(i) 
expressing the number of points included in the segment $P_i$ where the minimum distance is smaller and larger, respectively, 
than the threshold {\rm MinDistance} (lines 9 and 10). We determine whether or not we merge the segment $\bar{P}_i$ into the point 
cloud $P$ depending on the ratio between ${\rm far}(i)$ and ${\rm near}(i)$. 

\bigskip

\noindent
{\bf Algorithm 1}\ \ Merging method between two point clouds\\
\\
1. for $i \leftarrow 1:s$ \\
2. \hspace*{0.5cm} near$(i)=0$ \\
3. \hspace*{0.5cm} far$(i)=0$ \\
4. end for\\
5. for $j \leftarrow 1:m$ \\
6. \hspace*{0.5cm} $d \leftarrow {\rm min}(|\bar{p}_1 - p_j| , \cdots, |\bar{p}_m-p_j|)$ \\
7. \hspace*{0.5cm} $k \leftarrow {\rm argmin}(|\bar{p}_1 - p_j| , \cdots, |\bar{p}_m-p_j|)$ \\
8. \hspace*{0.5cm} $t \leftarrow {\rm SegmentNumber}(\bar{p}_k)$ \\
9. \hspace*{0.5cm} if $ d < {\rm MinDistance}$ then : ${\rm near}(t) \leftarrow {\rm near}(t) + 1$ \\
10. \hspace*{0.43cm} else : ${\rm far}(t) \leftarrow {\rm far}(t) + 1$ \\
11. end for\\
12. for $i \leftarrow 1:s$\\
13. \hspace*{0.43cm} if $\frac{{\rm far}(i)}{{\rm near}(i)} < {\rm Threshold}$, then $P \leftarrow {\rm Merge}(P, \bar{P}_i)$\\
14. end for

\bigskip

We further segment the merged point cloud. For each segment of point cloud, we calculate the distance between a point 
in the segment and the object which pose is estimated during the previous picking trial. If the distance is less than 
the threshold, we use the result of pose estimation during the previous pick. On the other hand, if the distance is larger 
than the threshold, we newly estimate the pose of an object by using two step algorithm using the CVFH and CRF estimation and 
the ICP algorithm. 

\section{Experiment}

\begin{figure}
	\centering
	\includegraphics[width=9cm]{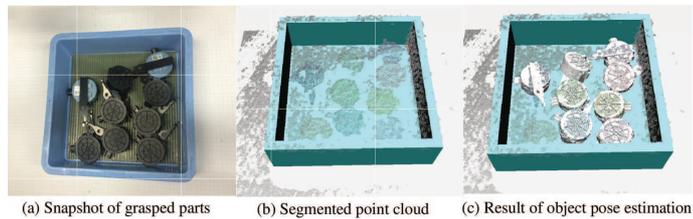}
	\caption{Estimation of objects' pose \label{fig:segmentation}}
\end{figure}

\begin{figure}
	\centering
	\includegraphics[width=9cm]{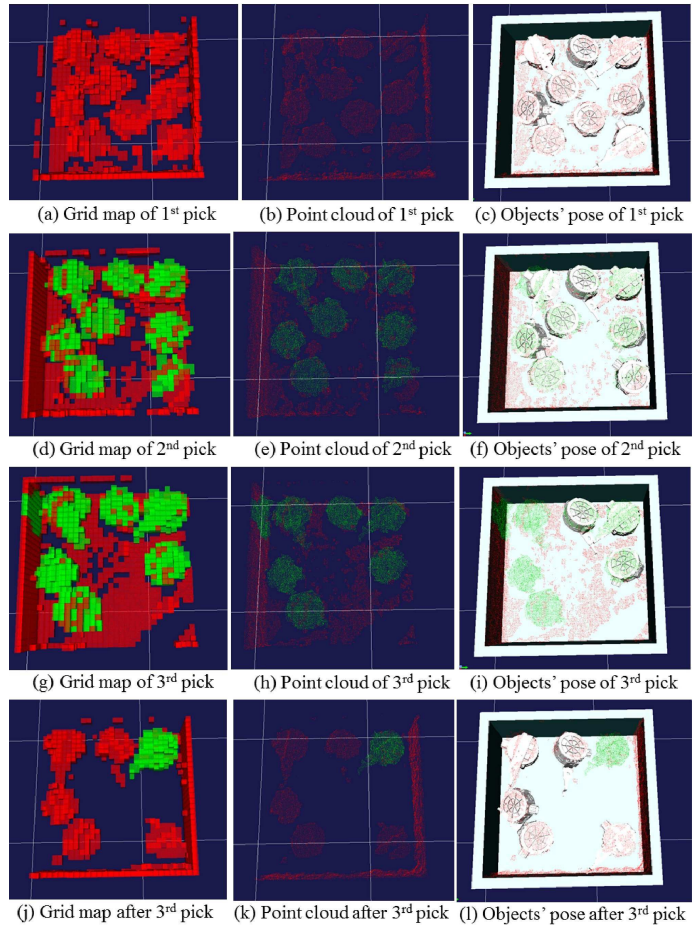}
	\caption{Grid cells of captured point cloud \label{fig:capture}}
\end{figure}

\begin{figure}
	\centering
	\includegraphics[width=8cm]{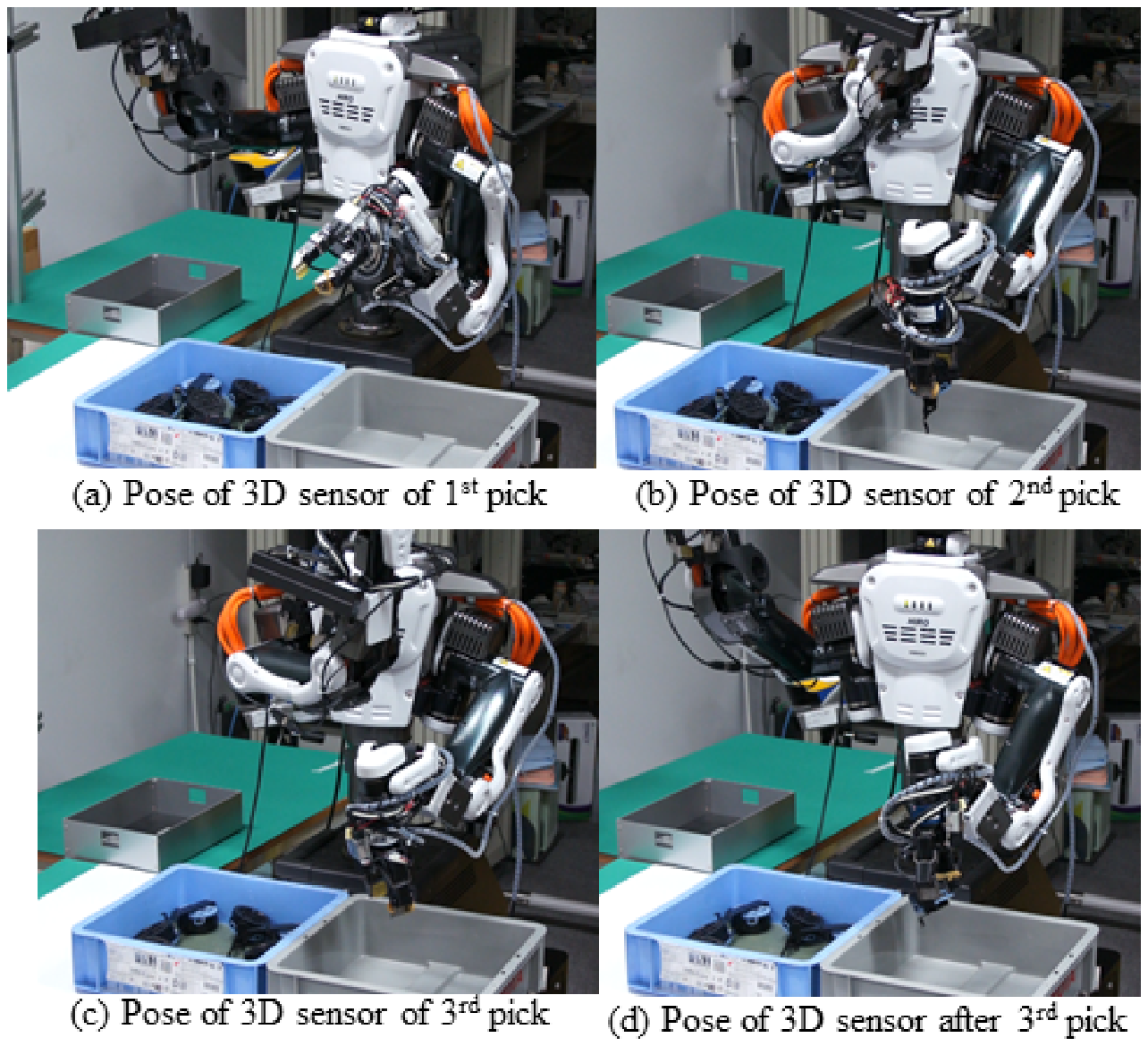}
	\caption{Pose of 3D sensor during a series of picking task \label{fig:camera}}
\end{figure}

We performed experiments on bin-picking. 
As shown in Fig. \ref{fig:segmentation}(a), we randomly placed nine objects in a box. 
We put nine objects close to each other such that the finger contacts 
a neighboring object when picking the target one. 
In the experiment, we performed the picking trial for three times. After the three times picking trial, we additionally 
captured the visual information. 
Fig. \ref{fig:capture} shows the grid cells of captured point cloud during a series of picking tasks 
where the red cells include the newly captured point cloud while the green cells include the previously 
captured point cloud. We can see that object recognition is performed only for the object where red cells are included. 
Fig. \ref{fig:camera} shows the pose of 3D vision sensor during a series of picking task by using the dual-arm industrial 
manipulator HiroNX. 

%\begin{figure}
%	\centering
%	\includegraphics[width=7cm]{graphics/result_grasp.eps}
%	\caption{Calculation result of finger swept volume and grasping posture \label{fig:grasp}}
%\end{figure}
%
%
%\begin{figure}
%	\centering
%	\includegraphics[width=8cm]{graphics/experiment.eps}
%	\caption{Overview of picking experiment \label{fig:experiment}}
%\end{figure}

\section{Conclusions}

In this paper, we discussed the visual recognition system for learning based randomized bin-picking. 
We first explained the view planning method to maximize the visibility of randomly stacked objects. 
Then, since randomized bin-picking usually estimates the pose of a number of 
objects, we relaxed the computational cost of the object pose detection by 
using the visual information on randomly stacked objects captured during the current picking task 
together with the visual information captured during the previous picking tasks. 
Through experimental results, we confirmed that the computational cost of the object recognition is reduced.  

Here, in our visual recognition of randomly stacked objects, we used the conventional Euclidian cluster based method 
to segment the stacked objects. Using more advanced method on segmentation is considered to be our future topic. 
Also, performing experiment for different shaped objects is also considered to be our future research topic.

\end{document}